\documentclass[11pt,a4paper]{article}
\usepackage[hyperref]{acl2020}
\usepackage{times}
\usepackage{latexsym}
\usepackage{multirow}

\usepackage[T1]{fontenc}
\usepackage{inconsolata}
\usepackage{amsmath,amsfonts,amssymb}
\usepackage{xcolor}
\usepackage{subfigure}
\usepackage{csquotes}
\usepackage{url}
\usepackage{graphicx}
\usepackage{framed}

\aclfinalcopy % Uncomment this line for the final submission
\renewenvironment{quote}{%
   \list{}{%
     \leftmargin0.3cm   % this is the adjusting screw
     \rightmargin\leftmargin
   }
   \item\relax
}
{\endlist}

\title{An Exploratory Study of Argumentative Writing by Young Students: A Transformer-based Approach}

\author{Debanjan Ghosh, Beata Beigman Klebanov, Yi Song \\
Educational Testing Service\\
{\tt {dghosh,bbeigmanklebanov,ysong}@ets.org}}
\begin{document}
\maketitle
\begin{abstract}
We present a computational exploration of argument critique writing by young students. Middle school students were asked to criticize an argument presented in the prompt, focusing on identifying and explaining the reasoning flaws. This task resembles an established college-level argument critique task. Lexical and discourse features that utilize detailed domain knowledge to identify critiques exist for the college task but do not perform well on the young students’ data.  Instead, transformer-based architecture (e.g., BERT) fine-tuned on a large corpus of critique essays from the college task performs much better (over 20\% improvement in F1 score). Analysis of the performance of various configurations of the system suggests that while children's writing does not exhibit the standard discourse structure of an argumentative essay, it does share basic local sequential structures with the more mature writers. 
\end{abstract}

\section{Introduction} \label{section:introduction}
%main task - examples
%cbal - examples
% model
% 

Argument and logic are essential in academic writing as they enhance the critical thinking capacities of students. Argumentation requires systematic reasoning and the skill of using relevant examples to craft a support for one's point of view \cite{walton1996argumentation}. In recent  times, the surge in AI-informed scoring systems has made it possible to assess writing skills using automated systems. Recent research suggests the possibility of argumentation-aware automated essay scoring systems \cite{stab-gurevych-2017-recognizing}. %An automated system that captures aspects of argumentation could %An automated system can evaluate essays (for example, argumentative essays for college and graduate school ad-missions such as., GRE, TOEFL, etc.) based on the strength or weakness of arguments.  
%[\BBK{{\bf Yi}, would you have a good citation or two for this? References are outside of the page limit, so we can put a couple.}]. \DG{DG: added}
% for researchers to develop argumentative writing tasks
%that can be evaluated through
%\cite{beigman_klebanov_detecting_2017}

%intro of cbal
%\BBK{This paragraph focuses on a brief review of arg mining lit and its main foci -- needs a bit more beef.} 
%\DG { DG: edited it}

Most of the current work on computational analysis of argumentative writing in educational context focuses on automatically identifying the argument structures (e.g., argument components and their relations) in the essays \cite{stab_parsing_2017,persing2016end,nguyen2016context} and by predicting essay scores from  features derived from the structures (e.g.,  the  number  of claims and premises and the  number  of  supported  claims) \cite{ghosh-etal-2016-coarse}. Related research has also addressed the problem of scoring a particular dimension of essay quality, such as relevance to the prompt \cite{persing2014modeling},  opinions and their targets \cite{farra-somasundaran-burstein:2015:bea}, argument strength \cite{persing2015modeling}, among others.

%Recent research has also measured the quality of such arguments by ranking the convincingness of the arguments \cite{habernal2016argument} or capturing how  argument structures in essays are correlated to human scores \cite{ghosh-etal-2016-coarse}. 

%\BBK{This paragraph looks at the assessment arg data that has been addressed in the lit} 
While argument mining literature has addressed the educational context, it has so far mainly focused on analyzing college-level writing. For instance, \newcite{nguyen2018argument}  investigated argument structures in TOEFL11 corpus \cite{blanchard2013toefl11}; \newcite{beigman_klebanov_detecting_2017} and \newcite{persing2015modeling} analyzed writing of university students; \newcite{stab-gurevych-2017-recognizing} used data from ``essayforum.com'', where college entrance examination  is the largest forum. Computational analysis of arguments in young students' writing has not yet been done, to the best of our knowledge. Writing quality in essays by young writers has been addressed \cite{deane2014,attali2008developmental,Attali_Burstein_2006}, but identification of arguments was not part of these studies.

%\DG{Stab-gur is from an online essay corpus. Others, i.e., persing is based on essays written by undergraduate students}

%terms of the type of argumentative writings the major focus has been on college-level admission essays: essays that students write to critique a given prompt \cite{beigman_klebanov_detecting_2017}. However, a well-argued essay demonstrating innovative ideas with timely examples (besides correct punctuation and spelling) can be useful learning goals \cite{graham2007meta} also for young students, e.g., middle school students. Research in the assessment of argumentation for middle school students is, however, is widely lacking [need to emphasize this; check the chapter from beata+nitin book].

%prompts resembling already existing tasks such as college-level argument critiquing writing tasks
In this paper, we present a novel learning-and-assessment context where middle school students were asked to criticize an argument presented in the prompt, focusing on identifying and explaining the reasoning flaws.  Using a relatively small pilot data collected for this task, our aim here is to automatically identify good argument critiques in the young students' writing, with the twin goals of (a) exploring the characteristics of young students' writing for this task, and (b) in view of potential scoring and feedback applications. We start with describing and exemplifying the data, as well as the argument critique annotation we performed on it (section~\ref{section:data}). Experiments and results are presented in section~\ref{section:method}, followed by a discussion in section~\ref{section:discussion}.

\section{Dataset and Annotation} \label{section:data}

\iffalse
\begin{figure}[t]
\centering
%\begin{framed}
%\includegraphics[width=5cm]{model1_1_new_bw.png}
\includegraphics[width=7.5cm]{cbal_picture.png}

%\end{framed}
\caption{Ban Ads Argument Critique Task (\cite{song2017examining}, p. 6)}
\label{figure:cbalprompt}
\end{figure}
\fi

\begin{table}
\centering
\small
%\begin{tabular}{ |l|p{10cm}| } 
\begin{tabular}{ |p{7.5cm}| }
\hline
Dear Editor,\\
\\
Advertising aimed at children under 12 should be allowed for  several reasons. \\
\\
First, one family in my neighbourhood sits down and watches TV together almost every evening. The whole family learns a lot, which shows that advertising for children is always a good thing because it brings families together. \\
\\
Second, research shows that children can't remember commercials well anyway, so they can't be doing kids any harm. \\
 \\
Finally, the arguments against advertising aren't very effective. Some countries banned ads because kids thought the ads were funny. But that's not a good reason. Think about it: the advertising industry spends billions of dollars a year on ads for children. They wouldn't spend all the money if the ads weren't doing some good. Let's not hurt children by stopping a good thing.\\ 
\\
If anyone doesn't like children's ads, the advertisers should just try to make them more interesting. The ads are allowed to be shown on TV, so they shouldn't be banned. \\
\hline

%\multicolumn{1}{|c|}{Platform} & {c}{Turn Type} & 
%\multicolumn{1}{c|}{Turn pairs} \\
\end{tabular}
\caption{The prompt of the argument critique task.} %(\cite{song2017examining}, p. 6).}
\label{table:cbalprompt}
\end{table}

 %\BBK{{\bf Yi}, if you could please draft this section, it would be great. Please describe the bigger picture of CBAL assessment and the argument critique piece of it more specifically. Ideally, the description of the task would also point out ways in which it is similar and different from GRE Arg. Please also describe the data collection that yielded the data we are working with -- when and where collected, how many kids. We want the reader to get a sense of the context from which the data comes. We also need the prompt (if it is OK to cite it) and examples of critiques and non-critiques. In the second part, please describe the annotation -- how the guidelines were adapted from the GRE annotation study (if they were), and kappa agreement on the distinction between generic and non-generic, if you have it. I think we want to emphasize that only reasonable critiques are annotated -- perhaps put an example of something that seems to be using critical language but does not quite make sense. This is an important distinction for the target community -- people mostly (though not always) look at structures, not whether the content makes sense.}

The data used in this study was collected as part of a pilot of a scenario-based assessment of argumentation skills with about 900 middle school students \cite{song2017examining}.\footnote{The data was collected under the ETS CBAL (Cognitively Based Assessment of, for, and as Learning) Initiative.} Students engaged in a sequence of steps in which they researched and reflected on whether advertising to children under the age of twelve should be banned. The test consists of four tasks; we use the responses to Task 3 in which students are asked to review a letter to the editor and evaluate problems in the letter's \emph{reasoning} or use of \emph{evidence} (see Table \ref{table:cbalprompt}). 

% (henceforth, ``Ban Ads''). % and, in the culminating task, write essays to present and support their positions on this issue. %The Ban Ads test consists of four tasks, and \newcite{song_learning_2016} analyzed students' responses in Task 3 in which they are asked to review a letter to the editor and evaluate arguments in the letter.

Students were expected to produce a written critique of the arguments, demonstrating their ability to identify and explain problems in the reasoning or use of evidence. For example, the first excerpt below shows a well-articulated critique of the hasty generalization problem in the prompt:
\iffalse
\begin{table}[h]
\centering
%\begin{tabular}{ |l|p{10cm}| } 
\begin{tabular}{ |p{7.5cm}| }
\hline
(1) Just because it brings one family together to learn does not mean that it will bring all families together to learn.\\
\hline
\end{tabular}
\label{table:cbalexample1}
\end{table}
\fi

\begin{quote}
 (1) Just because it brings one family together to learn does not mean that it will bring all families together to learn. 
 \end{quote}
 \begin{quote}
 (2) The first one about the family in your neighborhood is more like an opinion, not actual information from the article. 
 \end{quote}
 \begin{quote}
 (3) Their claims are badly writtin [sic] and have no good arguments. They need to support their claims with SOLID evidence and only claim arguments that can be undecicive [sic].
\end{quote}

However, many students had difficulty explaining the reasoning flaws clearly. In the second excerpt, the student thought that  an argument from the family in the neighborhood is not strong, but did not demonstrate an understanding of a weak generalization in his explanation.  
\iffalse
\begin{quote}
 (2) The first one about the family in your neighborhood is more like an opinion, not actual information from the article. 
\end{quote}
\fi
Other common problems included students summarizing the prompt  without criticizing, or providing a generic critique that does not adhere to the particulars of the prompt (excerpt (3)).

%The persons argument is based purely on opinion and is not backed up by any solid evidence. 
\iffalse
\begin{quote}
(3) Their claims are badly writtin [sic] and have no good arguments. They need to support their claims with SOLID evidence and only claim arguments that can be undecicive [sic]. \end{quote}
\fi
The goal of the argument critique annotation (described next) was to identify where in a response {\em good} critiques are made, such as the one in the first excerpt. %, in view of subsequent scoring and feedback. 

%See Figure \ref{figure:cbalprompt} for the essay prompt. For details regarding the design of the task please see \cite{song_learning_2016}. 

\paragraph{Annotation of Critiques:} 
We identified 11 valid critiques of the arguments in the letter. These critiques included: (1) overgeneralizing from a single example; (2) example irrelevant to the argument; (3) example misrepresenting what actually happened; (4) misrepresenting the goal of making advertisements; (5) misunderstanding the problem; (6) neglecting potential side effects of allowing advertising aimed at children; (7) making a wrong argument from sign; (8) argument contradicting authoritative evidence; (9) argument contradicting one's own experience; (10) making a circular argument; (11) making contradictory claims. All sentences containing any material belonging to a valid critique were marked and henceforth denoted as $Arg$; the rest are denoted as $NoArg$. Three annotators were employed to annotate the sentences to mark them as $Arg$/$NoArg$. We computed $\kappa$ between each pair of annotators based on the annotation of 50 essays. Inter-annotator agreement for this sentence-level  $Arg$/$NoArg$ classification for each pair of annotators was 0.714, 0.714, and 0.811, respectively resulting in an average $\kappa$ of 0.746.

\paragraph{Descriptive statistics:} 
We split the data into {\em training} (585 response critiques) and {\em test} (252 response critiques). The {\em training} partition has 2,220 sentences (515 $Arg$; 1,705 $NoArg$; average number of words per sentence is 11 (std = 8.03));  {\em test} contains 973 sentences. % \BBK{{\bf Debanjan}, I think we also need av/std of response length in sentences/words.}

%Observe that the training data is highly unbalanced with much more instances of $NoArg$.   

%intro of CBAL data
%As stated in the Introduction section, the training and test data are collected from  a new writing task for young students. This corpus consists of argumentative essays written by young students critiquing giving prompt. The essays ere annotated by [add].

\section{Experiments and Results} \label{section:method}

\subsection{Baseline}

In this writing task, young students were  asked to analyze the given prompt, focusing on identifying and explaining its reasoning flaws. This task is similar to a well-established task for college students previously discussed in the literature \citep{beigman_klebanov_detecting_2017}. Compared to the college task, the prompt for children appears to have more obvious reasoning errors. The tasks also differ in the types of responses they elicit. While the college task elicits a full essay-length response, the current critique task elicits a shorter, less formal response.
%with structural elements such as an introduction and conclusion

%song2017toward,

As our baseline, we evaluate the features that were reported as being effective for identifying argument critiques in the context of the college task. \newcite{beigman_klebanov_detecting_2017} described a logistic regression classifier with two types of features: 
\begin{itemize}
    \item features capturing discourse {\bf structure}, since it was found that argument critiques tended to occupy certain consistent discourse roles that are common in argumentative essays (such as the {\sc support}, rather than {\sc thesis} or {\sc background} roles), as well as have a tendency to participate in roles that receive a lot of elaboration, such as a {\sc support} sentence following or preceding another {\sc support} sentence, or a {\sc conclusion} sentence followed by another sentence in the same role.
    \item features  capturing {\bf content}, based on hybrid word and POS ngrams (see \newcite{beigman_klebanov_detecting_2017} for more  detail).

\end{itemize}

Table \ref{table:cbaldiscresults} shows the results, with each of the two subsets of features separately and together. Clearly, the classifier performs quite poorly for detecting $Arg$ sentences in children's data. Secondly, it seems that whatever performance is achieved is due to the content features, while the structural features fail to detect $Arg$. Thus, the well-organized nature of the mature writing, where essays have identifiable discourse elements such as {\sc thesis, main  claim, support, conclusion} \cite{burstein_finding_2003}, does not seem to carry over to young students' less formal writing. %It is possible, of course, that children's writing has a different discourse structure; it is also possible that young writers do not as yet organize their written discourse in a consistent fashion.

%\newcite{BeigmanEtAlContinuous2017}
%presents that discrete lexical features such as n-grams with parts-of-speech (POS) perform well in the case of same topic context (i.e., training within prompt) and discourse features, i.e., predicted discourse units such as Thesis, Background, Main-Point, Support, Conclusion, etc. \cite{burstein_finding_2003} perform best for different topic contexts (i.e., training across prompts).\footnote{For more description of the features please see \cite{BeigmanEtAlContinuous2017}.} We evaluated the discrete feature-bases models learned using a logistic regression classifier. However, both these sets of features, i.e., 1-3grams+POS and discourse features result in poor prediction of identifying arguments from young students' writings. Table \ref{table:cbaldiscresults} presents the performance of ngrams with POS (denoted as ``1-3gr ppos'') features with and without discourse relations (denoted as ``dr\_pn''). The F1 scores for $Arg$ is low and it is about 44.0\%.

\begin{table}
\centering
\small
%\begin{tabular}{ |l|p{10cm}| } 
\begin{tabular}{ p{2.2cm}p{1cm}p{0.8cm}p{0.8cm}p{0.8cm} }
\hline
%\multicolumn{1}{|c|}{Platform} & {c}{Turn Type} & 
%\multicolumn{1}{c|}{Turn pairs} \\
 Features & Category & Precision & Recall & F1 \\
\hline
%1-3gr ppos
\multirow{2}{*}{Content} & $NoArg$ &  0.851 &  0.946 &     0.896 \\
&  $Arg$ &  0.611 &  0.338 &     0.436 \\
\hline
\multirow{2}{*}{Structure} & $NoArg$ & 0.799 &  1.00 &     0.889 \\
&  $Arg$ &  0 &  0 &    0 \\
\hline
%dr\_pn+1-3gr ppos
\multirow{2}{*}{Structure + Content} & $NoArg$ &  0.852 &  0.940 &     0.894 \\
&  $Arg$ &  0.591 &  0.349 &     0.439 \\
\hline
\end{tabular}
\caption{Performance of baseline features. ``Structure" corresponds to the {\em dr$\_$pn} feature set, ``Content'' corresponds to the {\em 1-3gr ppos} feature set, both from \newcite{beigman_klebanov_detecting_2017}.}
\label{table:cbaldiscresults}
\end{table}

\subsection{Our system}

%the recent advancement in various NLP applications by the use of

As the training dataset is relatively small, we leverage pre-trained language models that are shown to be effective in various NLP applications.  Particularly, we focus  on BERT \cite{devlin2018bert}, a bi-directional transformer \cite{vaswani2017attention} based architecture that has produced excellent performance on argumentation tasks such as argument component and relation identification \cite{chakrabarty2019ampersand} and argument clustering \cite{reimers2019classification}. The BERT model is initially trained over  a  3.3  billion  word  English  corpus on two tasks: (1) given a sentence containing multiple masked words predict the identity of a particular masked word, and (2) given two sentences, predict whether they are adjacent. The BERT model exploits a multi-head attention operation to compute context-sensitive representations for each token in a sentence. During its training, a special token ``[CLS]'' is added to the  beginning of each training utterance. During evaluation, the learned representation for this ``[CLS]'' token is processed by an additional layer with nonlinear activation.  A standard pre-trained BERT model can be used for transfer learning when the model is ``fine-tuned'' during training, i.e., on the classification data of $Arg$ and $NoArg$ sentences (i.e., {\em training} partition) or by first fine-tuning the BERT language-model itself on a large unsupervised corpus from a partially relevant domain, such as a  corpus of writings from advanced students and then again fine-tuned on the classification data. In both the cases, BERT makes predictions via the ``[CLS]'' token.
%using a supervised learning objective.

%The BERT model exploits a multi-head attention operation to compute context-sensitive representations for each token in a sentence. 

\textbf{Fine-tuning on classification data}:
We first fine-tune a pre-trained BERT model (the ``bert-base-uncased'' version) with the $training$ data. During training the class weights are proportional to the numbers of $Arg$ and $NoArg$ instances. Unless stated otherwise we kept the following parameters throughout in the experiments:  we utilize a batch size of 16 instances, learning\_rate of 3e-5, warmup\_proportion 0.1, and the Adam optimizer. Hyperparameters were tuned for only five epochs. This experiment is denoted as BERT$_{bl}$ in Table \ref{table:cbalbertresults}. We observe that the F1 score for $Arg$ is 56\%,  resulting in a 12\% absolute improvement in F1 score over the structure+content features (Table \ref{table:cbaldiscresults}). This confirms that BERT is able to perform well even after fine-tuning with a relatively small training corpus with default parameters. 

 %statistical models based on discrete features described in Table \ref{table:cbaldiscresults}.
 
\begin{table}
\centering
\small
%\begin{tabular}{ |l|p{10cm}| } 
\begin{tabular}{ p{2cm}p{.8cm}p{1cm}p{1cm}p{0.8cm} }
\hline
%\multicolumn{1}{|c|}{Platform} & {c}{Turn Type} & 
%\multicolumn{1}{c|}{Turn pairs} \\
 Experiment & Category & Precision & Recall & F1 \\
\hline
\multirow{2}{*}{$BERT_{bl}$} & $NoArg$ &  0.884 &  0.913 & 0.898 \\
&  $Arg$ &  0.603 &  0.523 &     0.560 \\
\hline
\multirow{2}{*}{$BERT_{pair}$} & $NoArg$ &  0.892 &  0.934 &     0.913 \\
&  $Arg$ &  \textbf{0.681} &  0.556 &     0.612 \\
\hline
\multirow{2}{*}{$BERT_{bl+lm}$} & $NoArg$ &  0.907 &  0.898 &     0.902 \\
&  $Arg$ &  0.610 &  0.636 &     0.623 \\
\hline
\multirow{2}{*}{$BERT_{pair+lm}$} & $NoArg$ &  0.929 &	0.871 &	0.900\\
&  $Arg$ &  0.592 &	\textbf{0.740} &	\textbf{0.658} \\
\hline
\end{tabular}
\caption{Performance of BERT transformer, various configurations. Rows 1, 2 present results of BERT fine-tuning with $training$ data only; rows 3, 4 present the effect of additional language model fine-tuning. Highest scores are \textbf{bold}.}
\label{table:cbalbertresults}
\end{table}

In the next step, we re-utilize the same pre-trained BERT model while transforming the {\em training} instances to \emph{paired} sentence instances, where the first sentence is the candidate $Arg$ or $NoArg$ sentence and the second sentence of the pair is the immediate next sentence in the essay. For instance, for the first example in section 2, ``Just because \dots to learn'', now the instance also contains the subsequent sentence: 
\begin{quote}
    <Just because \dots to learn.>,<Second, children can't remember commercials anyway, so they can't be doing any harm," says the letter.>
\end{quote}

%[add an example from the set of examples earlier in the paper]. 
%\DG{Added.}
A special token ``FINAL\_SENTENCE'' is used when the candidate $Arg$ or $NoArg$ sentence is the last sentence in the essay. This modification of the data representation might help the BERT model for two reasons. First, pairing of the candidate sentence and the next one will encourage the model to more directly utilize the next sentence prediction task. Secondly, since multi-sentence same-discourse-role elaboration was found to be common in \newcite{beigman_klebanov_detecting_2017} data, BERT may exploit such sequential structures if they at all exist in our data. This is model BERT$_{pair}$ in Table \ref{table:cbalbertresults}.  With the paired-sentences transformation of the instances  the F1 improves to 61.2\%,  a boost of 5\%  over BERT$_{bl}$.       

%optimize on the

\textbf{Fine-tuning with a large essay corpus}: 
It has been shown in related research \cite{chakrabarty2019ampersand} that transfer learning  by  fine-tuning on a domain-specific corpus using a supervised learning objective can boost  performance.  We used a large proprietary corpus of college-level argument critique essays similar to those analyzed by \newcite{beigman_klebanov_detecting_2017}. This corpus consists of 351,363 unannotated essays, where an average  essay contains 16 sentences, resulting in a corpus of 5.64 million sentences. We fine-tune the pre-trained BERT language model on this large corpus for five epochs and then again fine-tune it with the $training$ partition (BERT$_{bl+lm}$). Likewise,  BERT$_{pair+lm}$ represents the model after pre-trained BERT language model is fine-tuned with the large corpus and then again fine-tuned with the paired instances of $training$.   
% \BBK{This isn't very clear -- is the GRE fine-tuning done on pairs as well? If not, it is not quite parallel to CBAL fine-tuning described above...}. 
% \DG { edited a bit. please see.}
We observe that fine-tuning the language model improves F1 to 62.3\% whereas BERT$_{pair+lm}$ results in the highest F1 of 65.8\%, around 5\% higher than  BERT$_{pair}$ and over 20\% higher than the feature-based model.

\iffalse
\begin{table}
\centering
\small
%\begin{tabular}{ |l|p{10cm}| } 
\begin{tabular}{ p{2cm}p{1cm}p{0.8cm}p{0.8cm}p{0.8cm} }
\hline
%\multicolumn{1}{|c|}{Platform} & {c}{Turn Type} & 
%\multicolumn{1}{c|}{Turn pairs} \\
 Experiment & Category & P & R & F1 \\
\hline
\multirow{2}{*}{BERT_{bl+lm}} & $NoArg$ &  0.907 &  0.898 &     0.902 \\
&  $Arg$ &  0.610 &  0.636 &     0.623 \\
\hline
\multirow{2}{*}{BERT_{pair+lm}} & $NoArg$ &  0.929 &	0.871 &	0.900\\
&  $Arg$ &  0.592 &	0.740 &	0.658 \\
\hline
\end{tabular}
\caption{Performance after additionally fine-tuning the language model with a large unannotated essay corpus.}
\label{table:cbalbertlmresults}
\end{table}
\fi
%\input{results}
\section{Discussion} \label{section:discussion}

The difference in F1  between BERT$_{bl}$, BERT$_{bl+lm}$, and BERT$_{pairs+lm}$ is almost exclusively in recall -- they have comparable precision at about 0.6, with recall of 0.52, 0.64, and 0.74, respectively. Partitioning out 10\% of the {\em training} data for a {\em development} set, we found that BERT$_{bl+lm}$  detected 13 more $Arg$ sentences than BERT$_{bl}$ in the development data. These fell into two {\bf sequential} patterns: (a) the sentence is followed by another that further develops the  critique (7 cases) --  see excerpts (4) and (5) below; (b) the sentence is the final sentence in the response (6 cases);  excerpt (6).

\begin{quote}
(4) They werent made to be appealing to adults.	They only need kids to want the product, and beg their parents for it.
\end{quote}
\begin{quote}
(5) Finally, is spending billions of dollars on something that has no point a good thing?	There are many arguements  that all this money is just going to waste, and it could be used on more important things.	
\end{quote}

%\begin{quote}
%The ads are only putting a positive look on their product.	not a childs' health.
%\end{quote}

\begin{quote}
(6) I say this because in an article I found out that children do remember advertisements that they have seen before. \end{quote}

%\begin{quote}
%if they cant remeber the commercials than why is the advertising industrie spending billions of dollars on the commercilas.
%\end{quote}

%\begin{quote}
%(4) Peer pressure is also incorperated into advertisements.
%\end{quote}

Our interpretation of this finding is that BERT$_{bl+lm}$ captured organizational elements in children's writing that  {\em are} similar to adult patterns. \newcite{beigman_klebanov_detecting_2017} found that adult writers often reiterate a previously stated critique in an extended {\sc conclusion} and spread critiques  across  consecutive {\sc support} sentences.  Thus, even though alignment of critiques with ``standard" discourse elements such as {\sc conclusion} and {\sc support} is not recognizable in children's writing (as witnessed by the failure of the structural  features to detect  critiques), some {\bf basic local sequential patterns} do exist, and they are sufficiently similar to the ones in adult writing that a system with its language model tuned on adult critique writing can capitalize on this knowledge.

Interestingly, BERT$_{pairs}$ learned similar sequential patterns -- indeed 7 of the 13 sentences gained by BERT$_{bl+lm
}$ over BERT$_{bl}$ are also recalled by BERT$_{pairs}$. This further reinforces the conclusion that young writers exhibit certain local sequential patterns of discourse organization that they share with mature argument critique writers.

% (emphasize adult-child contrast;  rhetorical question):

  % Indeed, in examples 1 and 2, the first sentence is a fairly sophisticated setup to reinforce the point made in the second. Such responses might not be frequent enough in the small children's data for effective learning, but, having been tuned to expect certain patterns in good critiques, the system can capitalize on the knowledge.
  
%while BERT$_{bl}$  shows recall of 0.52 vs BERT$_{pairs+lm}$ much better recall of 0.74. We therefore investigated $Arg$ instances that were correctly recalled by the latter model but were missed by the former one in order to get insight into the kind of knowledge that the improved model has acquired from both the paired sentences and the language model tuning on the large corpus of argument critique essays from adult writers.

% we used the remaining $training$ instances to BERT$_{bl}$ and BERT$_{lm}$ models, to examine the effect of language model adaptation.  The latter
\section{Conclusion and Future Work} \label{section:conclusion}

%We showed that fine-tuning the language model of BERT achieves the best performance in detecting argument critique from writings of middle school students. While traditional discourse and content features perform poorly, we observe BERT does capture the fact that young students do share some basic local sequential structures with the more mature writers. In future, we want to expand the number of argument prompts to investigate the cross-prompt behaviors of the transformer models.

We present a computational exploration of argument critiques written by middle school children. A feature set designed for {\em college-level} critique writing has poor recall of critiques when trained on children's data; a pre-trained BERT model fine-tuned on children's data does better by 18\%. When BERT's language model is additionally fine-tuned on a large corpus of {\em college} critique essays, recall improves by further 20\%, suggesting the existence of {\em  some} similarity between young and mature writers. Performance analysis suggests that BERT capitalized on certain  sequential patterns in critique writing; a larger study examining patterns of argumentation in children's data is needed to confirm the hypothesis. In future, we plan to fine-tune our models on auxiliary dataset, such as the convincing argument dataset from \newcite{habernal2016argument}. 
%In addition, using only the high scoring essays for LM is also a plan.
%intermediate task transfer 

\bibliography{acl2019}
\bibliographystyle{acl_natbib}

\end{document}